\documentclass[conference]{IEEEtran}
\IEEEoverridecommandlockouts
\usepackage{cite}
\usepackage{amsmath,amssymb,amsfonts}
\usepackage{algorithmic}
\usepackage{graphicx}
\usepackage{textcomp}
\usepackage{xcolor}

\def\BibTeX{{\rm B\kern-.05em{\sc i\kern-.025em b}\kern-.08em
    T\kern-.1667em\lower.7ex\hbox{E}\kern-.125emX}}

\makeatletter 
\newcommand{\linebreakand}{%
  \end{@IEEEauthorhalign}
  \hfill\mbox{}\par
  \mbox{}\hfill\begin{@IEEEauthorhalign}
}
\makeatother 

\newcommand{\source}{\mathcal{S}}

\begin{document}

\title{Towards Holistic Disease Risk Prediction \\using Small Language Models
}
\author{\thanks{$\dagger$ These authors contributed equally to this work.} \IEEEauthorblockN{1\textsuperscript{st} Liv Bj\"{o}rkdahl$^\dagger$}
\IEEEauthorblockA{\textit{AI Sweden}\\
Gothenburg, Sweden \\
liv.m.bjorkdahl@gmail.com}
\and
\IEEEauthorblockN{2\textsuperscript{nd} Oskar Pauli$^\dagger$}
\IEEEauthorblockA{\textit{AI Sweden} \\
Gothenburg, Sweden \\
oskar@pauli.nu}
\and
\IEEEauthorblockN{3\textsuperscript{rd} Johan \"Ostman}
\IEEEauthorblockA{\textit{AI Sweden}\\
Gothenburg, Sweden \\
johan.ostman@ai.se}
\and
\IEEEauthorblockN{4\textsuperscript{th} Chiara Ceccobello}
\IEEEauthorblockA{
\textit{AI Sweden}\\
Gothenburg, Sweden \\
chiara.ceccobello@ai.se}
\and
\linebreakand
\IEEEauthorblockN{5\textsuperscript{th} Sara Lundell}
\IEEEauthorblockA{
\textit{Sahlgrenska University Hospital}\\
Gothenburg, Sweden \\
sara.lundell@vgregion.se}
\and
\IEEEauthorblockN{6\textsuperscript{th} Magnus Kjellberg}
\IEEEauthorblockA{
\textit{Sahlgrenska University Hospital}\\
Gothenburg, Sweden \\
magnus.kjellberg@vgregion.se}
}

\maketitle
\begin{abstract}
Data in the healthcare domain arise from a variety of sources and modalities, such as x-ray images, continuous measurements, and clinical notes. Medical practitioners integrate these diverse data types daily to make informed and accurate decisions. With recent advancements in language models capable of handling multimodal data, it is a logical progression to apply these models to the healthcare sector. In this work, we introduce a framework that connects small language models to multiple data sources, aiming to predict the risk of various diseases simultaneously. Our experiments encompass 12 different tasks within a multitask learning setup. Although our approach does not surpass state-of-the-art methods specialized for single tasks, it demonstrates competitive performance and underscores the potential of small language models for multimodal reasoning in healthcare.
\end{abstract}


\section{Introduction}

Effective healthcare systems are crucial for managing public health, addressing disease outbreaks, and providing timely medical interventions. By leveraging advanced machine learning (ML) techniques, it is possible to enhance patient care, streamline administrative processes, and support medical research~\cite{davenport2019potential}.
Healthcare data is inherently multimodal and heterogeneous, encompassing various types of information such as clinical notes, medical imaging, lab event data, and continuous measurements. For instance, a patient’s medical record may include textual descriptions of symptoms, X-ray images, demographic information, and continuous measurements of vital signs such as blood pressure or heart rate. This diversity in data types, i.e. modalities, poses significant challenges for analysis due to the varying formats, levels of detail, and degrees of completeness across different modalities. 
Furthermore, patients might be affected by several diseases simultaneously, demanding a medical practitioner to understand and piece together a complex set of evidence in order to formulate an effective diagnosis.

Multimodal models are expected to outperform models trained on a single modality due to an improved quality of the latent space representation~\cite{huang2021makes}.
Recently, this has been shown empirically in the healthcare setting, see, e.g.,~\cite{Krones2024ReviewOM, soenksen2022integrated}, for a plethora of different downstream tasks. 
In these works, the latent representation is obtained via joint fusion where modality-specific pretrained encoders are used to create an aggregated feature representation of the multiple modalities.
Alternatives to the joint-fusion approach has recently surfaced to address the missing-modality issue, relying on sequential fusion where modalities are encoded using sequential models like Long Short-term Memory~\cite{hayat2022medfuse}.

Recently, large languge models (LLMs) have been introduced to the multimodal scene for healthcare.
To enable multimodal reasoning, it is common to freeze the LLM and either train projectors that map the different modalities to the token space~\cite{belyaeva2023multimodal} or to embed the input data into the prompt~\cite{belyaeva2023multimodal, kim2024health, shoham2024cpllmclinicalpredictionlarge}. 
In particular, in~\cite{belyaeva2023multimodal}, the disease risk predictions significantly improved when exposed to multiple modalities rather than single ones. 
However, the challenges imposed by LLMs, or multimodal LLMs, in terms of infrastructure, energy costs and computing time makes them impractical for many healthcare providers to deploy and use.

To overcome these issues associated with LLMs, smaller versions of pre-trained LLMs, so-called small language models (SLMs), are emerging by means of knowledge distillation \cite{hinton2015distilling} and quantization~\cite{wan2023efficient}.
Noticeably, SLMs of just a few billion parameters show performances compared to even much large models, see~\cite{abdin2024phi,imp2024,yang2024advancing}.

In this study, we explore how SLMs, in particular Gemma-2B~\cite{gemmateam2024gemma} and Phi-3-mini-4k~\cite{abdin2024phi}, can be used to process multimodal inputs (time series data, text and visual data) to perform predictions for various health related tasks. Following the framework described in \cite{soenksen2022integrated}, we train one single model to perform length of stay-prediction, 48h mortality-prediction, and 10 different prediction tasks associated with chest pathology diagnosis, namely, fracture, lung, lesion, enlarged cardiomediastinum, consolidation, pneumonia, atelectasis, lung opacity, pneumothorax, edema, and cardiomegaly.
In particular, our main contributions are:
\begin{itemize}
    \item We provide a scalable model-agnostic method to perform multi-disease risk prediction using SLMs. Our method is able to handle arbitrary modalities and is designed to account for imbalanced data, something often present in healthcare data. Moreover, the code is open-sourced\footnote{https://anonymous.4open.science/r/UNI-MLM-Unified-Multimodal-Language-Models-for-Multitask-Diagnosis-F930.}.
    \item We demonstrate our method using Gemma-2B and Phi-3-mini-4k and benchmark against state-of-the-art methods, tailored for a single task.
\end{itemize}

\begin{figure*}[t]
\centering
\includegraphics[width=0.8\textwidth]{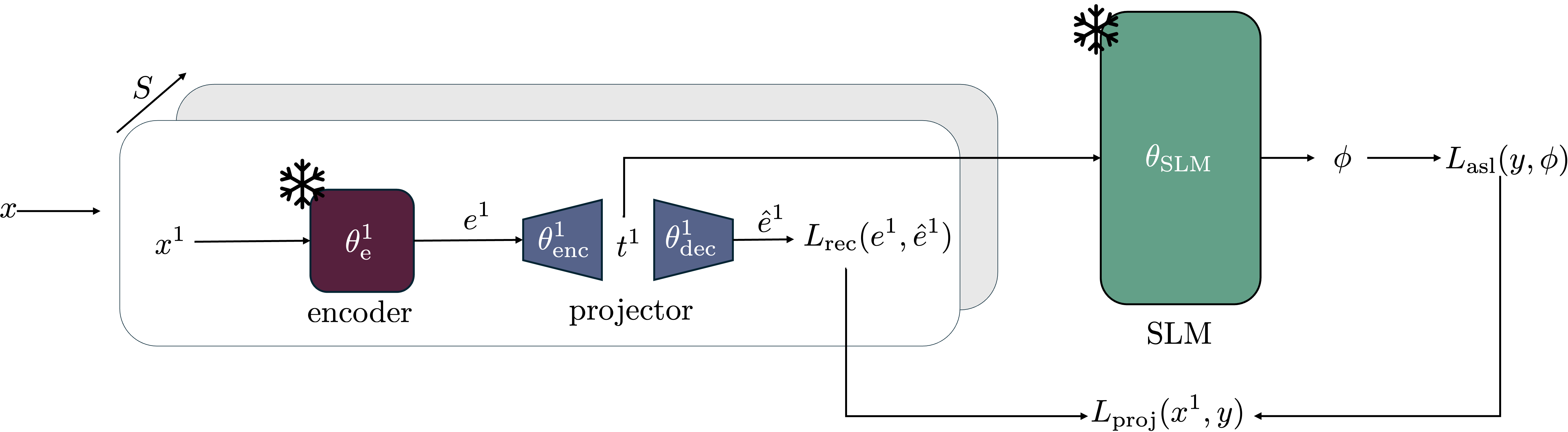}
\label{fig:method}
\caption{Joint training of projectors for $x$ containing $S$ sources based on the asymmetric loss~\eqref{eq:asl}. The pipeline shown for $x^1$ is replicated for all $x^s$, $s\in [S]$. The snowflake symbolizes that model weights are frozen.}
\end{figure*}

\section{Method}

\subsection{Multimodal Data Fusion}
We consider a setting where data is collected from multiple sources, e.g., notes, lab measurements, or screenings, with potentially different modalities, e.g., image, text, tabular, or time series.
We index the different sources from $1$ to $S$ and denote by $\source_s$ the $s$th source pertaining to a given modality.
Each source is assumed to generate data points $x_{i}^s$, denoted by $x_i^s\sim \mathcal{S}_s$, where $i\in [n]$ and $s\in[S]$ with $n$ being the number of data points.
Our goal is to build a model with the ability to jointly reason around the different sources and their corresponding modalities.
To achieve joint reasoning across modalities, the different sources must be fused together.

Data fusion has traditionally been approached by early, joint, or late fusion~\cite{huang2020fusion}.
In early fusion, the data generated from the different sources are concatenated at the input level, i.e., $x_i = \{ x_{i}^s \}_{s=1}^S$, and used to train the the model. 
However, as the semantics of different modalities may differ significantly, early fusion is not commonly used~\cite{huang2020fusion}.
Joint fusion attempts to map the different modalities into the same semantic setting by first creating latent representations, i.e., embeddings, for each source after which the embeddings are concatenated and used as input for model training.
Finally, late fusion maps the different modalities to separate output scores that are aggregated to yield the final decision.

In this work, we adopt a joint-fusion approach, that is, for each source $s\in[S]$, we consider a modality-dependent mapping $\theta^s:\mathcal{X}^{s} \rightarrow \mathcal{X}$ where $\mathcal{X}^{s}$ denotes the input space of the data generated from source $s$ and $\mathcal{X}$ is the  embedding space that all sources are mapped into. 
As outlined in Section~\ref{sec:oae}, this mapping is achieved in two stages, i.e., $\theta^s = \theta_{\mathrm{enc}}^s \circ\theta_\mathrm{e}^s$, where the source is first mapped into a modality-dependent intermediate latent space before being mapped to the joint embedding space.

\subsection{Multilabel classification}\label{sec:multilabel}

In the healthcare setting, a given patient may experience multiple diseases simultaneously, i.e., disease diagnosis is inherently multi-label~\cite{tsoumakas2007multi}. 
Let $K$ denote the number of different classes, i.e., diseases, to predict.
We denote by $y_{ik}$ the ground-truth label of the $k$th class of the $i$th data record, $k\in[K]$ and $i\in[n]$, and let $y_i = [y_{i1}, \dots, y_{iK}]$ denote the concatenation of ground-truth labels for the different classes. 
Although, in general, there may be multiple labels per class, herein we consider binary labels.
Moreover, patients often have labels for only a subset of the available classes or have labels indicating that the diagnosis is inconclusive.
To account for this, we let $y_{ik}\in\{\mathrm{u}, 0,1\}$ where $\mathrm{u}$ represents the union of a missing label and the inconclusive label.

As our goal is to train a single classifier $\theta$ to predict each of the $K$ classes, the loss function must account for all the different class predictions.
Taking into account that class imbalances are common in multi-label problems, and in healthcare specifically~\cite{soenksen2022integrated}, we next present two loss functions: the class-averaged weighted binary cross-entropy~\cite{cui2019class} and an asymmetric focal loss~\cite{ridnik2021asymmetric}.
Ignoring the sample index $i$, let $\theta$ be a classifier whose output confidence over the $K$ different classes is denoted by $\phi\in\mathbb{R}^K$.
We express a general loss function for our binary multi-label problem as~\cite{ridnik2021asymmetric}
\begin{equation}
\label{eq:loss}
    L(y, \phi) = -\sum_{k=1}^K y_k L_+(y_k, \phi_k) + (1-y_k) L_-(y_k, \phi_k).
\end{equation}

The first loss function we consider is the class-averaged weighted binary cross-entropy, denoted as $L_{\mathrm{avg}}$, where
\begin{align}
\label{eq:wbce}
\begin{split}
    L_+(y_k, \phi_k) &=  w_{k}^{\mathrm{pos}}\log(\phi_{k}) \\
    L_-(y_k, \phi_k) &=  w_{k}^{\mathrm{neg}}\log(1-\phi_k)
\end{split}
\end{align}
where $w_{k}^{\mathrm{i}}$ denotes the inverse class frequencies given as $w_{k,i} = n/(2K\sum_{p=1}^n \mathbf{1}\{y_{pk}=1\}$ with $\mathbf{1}\{E\}$ denoting the indicator function of an event $E$.
The idea in~\eqref{eq:wbce} is to focus more on the less prevalent classes.

A risk with~\eqref{eq:wbce}, due to the class average, is that a poor performance on a given class may be compensated by a strong performance on another. 
Furthermore, under significant class imbalance, the weights in~\eqref{eq:wbce} may not suffice to focus on the positive samples.
To alleviate these issues, we also consider an asymmetric loss~\cite{ridnik2021asymmetric}, denoted by $L_{\mathrm{asl}}$, given as
\begin{align}
\label{eq:asl}
\begin{split}
    L_+(y_k, \phi_k) &= (1-\phi_k)\log(\phi_{k})\\
    L_-(y_k, \phi_k) &= (p_{m})^{\gamma_-} \log(1-p_m)
\end{split}
\end{align}
where $p_{m}=\max\{\phi_k-m,0\}$ with $m\in[0,1]$. 
The main idea of~\eqref{eq:asl} is to emphasize the loss of misclassified positive labels, via $\gamma_-$, and to neglect negative samples that are easy to classify, i.e., $\phi_k\leq m$.

We obtain the final loss functions by considering only the classes with relevant labels, i.e.,
\begin{equation}
    \label{eq:loss_both}
    L_j(y, \phi) = -\sum_{k=1}^K L_{j}(y_k, \phi_k) \mathbf{1}\{ y_k \neq \mathrm{u} \}
\end{equation}
for $j\in\{ \mathrm{avg}, \mathrm{asl}\}$.

\subsection{Multimodal multilabel classification using SLMs}\label{sec:oae}

From the multimodal fusion, we have access to feature embeddings for each of the sources that can be used as input for model training.
In this work, we adopt the same methodology as~\cite{Moon2023AnyMALAE, belyaeva2023multimodal} where the capabilities of the pretrained LLM is to be leveraged by learning how to project the embeddings into the token space of the LLM.
Hence, the weights of the LLM are frozen and the projectors are trained by means of backpropagation leveraging the output prediction of the LLM.

In this work, for a given source $s\in[S]$, we consider an off-the-shelf encoder $\theta_{\mathrm{e}}^s$, pertaining to the modality at hand, to create source feature embeddings. 
However, for the embeddings to be used as input to a language model, they must be further projected into to the, often larger, token space. 
Since projecting an embedding into a larger space will risk to dilute the signal, we must ensure to maintain the signal while maintaining the discriminative properties of the embedding.
For this reason, we employ a regularized overcomplete autoencoder~\cite{bourlard2022autoencoders} as projector modules, i.e., an autoencoder with dimension of the bottleneck layer larger than the input and output dimensions.
We denote the encoder and decoder of the autoencoder by $\theta_{\mathrm{enc}}^s$ and $\theta_{\mathrm{dec}}^s$, respectively.
For source $s\in[S]$, the loss function of the projector is given as
\begin{equation}
    \label{eq:reg_ae}
    L_{\mathrm{proj}}(x^s, y) = L_{\mathrm{rec}}(e^s,\hat{e}^s) + \beta L_j(y, \phi)
\end{equation}
where $e^s=\theta_e(x^s)$ denotes the embedding of the data from source $s$, i.e., $x^s$, $\hat{e}^s=\theta_{\mathrm{dec}}(\theta_{\mathrm{enc}}(e^s))$ denotes the reconstruction of $e^s$, $L_{\mathrm{rec}}$ is the reconstruction loss, $j\in\{\mathrm{avg}, \mathrm{asl}\}$,  $\beta\geq 0$, and $\phi\in\mathbb{R}^K$ is the confidence prediction from the SLM based on the projection $t^s = \theta_{\mathrm{enc}}(e^s)$ in token space.
Note that $\phi$ is obtained by extracting the confidence from $K$ designated entries in the vocabulary, chosen at random before the training.

The projectors of the different sources can be trained jointly or in isolation.
In isolation, the projectors are trained on their individual inputs. 
That is, for the projector corresponding to source $s$, a data point $x^s \sim \mathcal{S}_s$ is mapped to an embedding $e^s$ which is projected into the token space of the SLM, i.e., $t^s = \theta_{\mathrm{enc}}(e^s)$.
The token $t^s$ is then used as input to the SLM to yield a prediction, i.e.,  $\phi = \theta_{\mathrm{SLM}}(t^s)$, to be used in the loss~\eqref{eq:loss_both}.

In isolated training, the projectors do not attempt to jointly create discriminative projections into token space. 
Therefore, we also consider joint training of the projector modules where the projectors are simultaneously mapping the source feature embeddings into token space, i.e., $t_i=\theta_{\mathrm{e}}^i$, after which the tokens are concatenated and fed in sequence to the SLM.
As the SLM is trained for next-token prediction, we obtain $S$ logit vectors as outputs, i.e., $\phi_s$ where $s\in[S]$, .
To create the final logits for prediction to be used in~\eqref{eq:loss_both}, we average the logit vectors, i.e., $\phi_i=\frac{1}{S}\sum_{s=1}^S \phi_{s}$, and threshold the resulting logit values to reach label predictions.
The procedure of jointly training the projectors for the different sources is outlined in Fig.~1.
We remark that both the encoders $\theta_\mathrm{e}^s$ and the SLM $\theta_{\mathrm{SLM}}$ are frozen during training whereas the projector modules are updated according to~\eqref{eq:loss_both}.



\section{Experiments}

\subsection{Data}\label{sec:data}
\begin{figure}[t]
\centering
\includegraphics[width=0.5\textwidth]{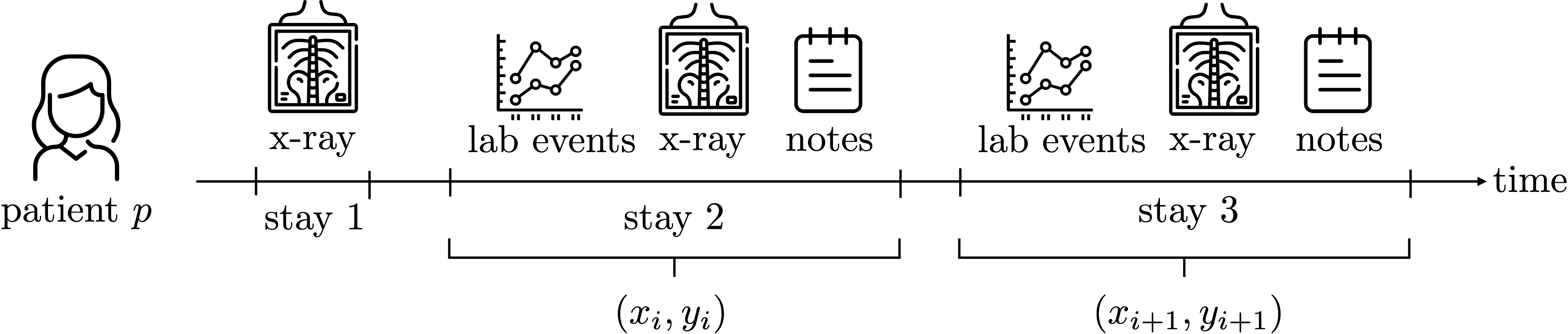}
\label{fig:data}
\caption{Data extraction process for a given patient $p$.}
\end{figure}
\begin{table}[t]
\label{tab:label_distr}
\caption{Total number of samples, number of negative samples, and number of positive samples for each label.}
    \centering
    \begin{tabular}{|l|l|r|r|}
        \hline
        & \# samples & neg & pos \\
        \hline
        Fracture & 1612 & 85 & 1527 \\
        Lung Lesion & 16111 & 100 & 1511 \\
        Enlarged CM & 6614 & 1831 & 4783 \\
        Consolidation & 9747 & 1701 & 8046 \\
        Pneumonia & 14684 & 6539 & 8145 \\
        Atelectasis & 30274 & 808 & 29466 \\
        Lung Opacity & 29540 & 1107 & 28433 \\
        Pneumothorax & 34171 & 27806 & 6365 \\
        Edema & 30713 & 11496 & 19217 \\
        Cardiomegaly & 34832 & 7072 & 27760 \\
        Length of stay & 90811 & 82323 & 8488 \\
        48h Mortality & 90811 & 88581 & 2230 \\
        \hline
        \multicolumn{4}{|c|}{Total dataset size: 90811} \\
        \hline
    \end{tabular}
\end{table}

We consider the MIMIC-IV v2.2, an anonymized dataset containing medical data from patients admitted to the intensity care unit (ICU) at the Beth Israel Deaconess Medical Center, MA, U.S., between 2008-2019~\cite{johnson2023mimic}. 
In particular, we consider data including critical care data~\cite{johnson2023mimic}, chest x-ray images~\cite{johnson2019mimic}, and  discharge summaries~\cite{johnson_notes23}.
Each of these data are indexed by patient ID and hospital stay, making it possible to link the same patient between multiple datasets.
In total, the dataset contains 26359 unique patient IDs.

In this work, we follow the approach in~\cite{soenksen2022integrated}  to extract data from procedure, lab and chart events (time series), chest x-ray screenings (images), and radiology notes (text). 
In particular, for each patient, we extract all different stays at the ICU and filter out the stays that do not include all the sources, i.e., stays lacking any of the sources are not added to the multimodal dataset. 
Note that a patient may have several stays at the hospital, hence, a single patient may have multiple data entries in the dataset, as can be seen in Fig.~2.
This step resulted in 90811 data records coming from 14854 unique patient IDs.

Next, for each ICU stay, we extract ten labels associated with tasks linked to chest pathology prediction (fracture, lesion, enlarged cardio mediastinum (CM), consolidation, pneumonia, atelectasis, lung opacity, pneumothrax, edema and cardiomegaly) along with labels for prediction of the length of stay (more or less than 48h) and 48h mortality prediction, resulting in a label vector consisting of 12 classes.
The resulting number of labels, positives, and negative samples of each class is shown in Table~\ref{tab:label_distr}.
As can be seen, the different classes are highly imbalanced and varies significantly in data quantity. 
For example, there are 90811 entries for length of stay prediction but only 1612 entries for fracture with imbalance ratio, i.e., positives/negatives, being 0.10 and 17.96, respectively.
This highlights the importance of treating the two classes differently via an asymmetric loss, as presented in Section~\ref{sec:multilabel}.

Next, we explain how data is extracted from the different sources and how embeddings are obtained.
In particular, we outline six different sources, involving three different modalities, yielding a data point $x$ and define the corresponding encoders $\theta_\mathrm{e}^s$, $s\in[6]$, to map $x$ into embeddings $e$.

\subsubsection{Procedure, lab, and chart events}
\label{sec:timeseries}
The sources corresponding to procedure, lab, and chart events yield 10, 22, and 9 time-series, respectively. 
To create embeddings, we consider, as an encoder, a mapping of each time-series onto 11 different quantities, i.e., the mean, variance, minimum, maximum, average difference, average absolute difference, maximal difference, sum of absolute difference, difference between first and last recordings, number of peaks (thresholded using median), and trend of the time series.
We then group the quantities pertaining to each time-series within each source and normalize the groups, resulting in $e^{\mathrm{proc}} \in \mathbb{R}^{110}$, $e^{\mathrm{lab}}\in\mathbb{R}^{242}$, and $e^{\mathrm{chart}}\in\mathbb{R}^{99}$.

\subsubsection{Images}
Consider a given stay resulting in $J$ x-ray screenings, i.e., (time, image) pairs $(t_j, x_j^{\mathrm{xr}})$, $j\in[J]$, where $t_j$ denotes the time of the screening.
For each image $x_j^{\mathrm{xr}}$, we create an embedding by processing the image, of size 224x224, via a Densenet121-Res224-CheX model, an adaption of DenseNet~\cite{huang2017densely}, of depth 121, pretrained on the CheXpert dataset~\cite{chexpert19}, resulting in $e_j^{\mathrm{xr}}\in\mathbb{R}^{1024}$.

For the x-ray screenings, we consider two sources: the most recent screening and the aggregate of all screenings for the duration of the stay.
In the former case, we obtain the embedding as $e^{\mathrm{xr}} =  e^{\mathrm{xr}}_{j'}$ where $j'=\arg\max_j\{t_j\}$.
To also use the older x-ray screenings, we perform a weighted average over the embeddings as $e^{\mathrm{axr}} = \sum_{j} w_j e_{j}^{\mathrm{xr}}$ with $w_j = (t_j - \min_j{t_j})/\max_j t_j$ for $j\in [J]$, again resulting in an embedding of size 1024.

\subsubsection{Text}
For each datapoint, we extract the radiology notes $x^{\mathrm{txt}}$ and employ Bio-BERT~\cite{lee2020biobert}, a BERT model~\cite{bert19} fine-tuned on text from MIMIC-III v1.4~\cite{johnson2016mimic}, to generate an embedding.
As the context window of Bio-BERT is 512 tokens, whenever $x^{\mathrm{txt}}$ exceeds this number, we divide the text into patches of 512 tokens and average the resulting embeddings.
This procedure yields an embedding $e^{\mathrm{txt}}\in\mathbb{R}^{768}$.

\subsection{Setup}

The procedure in Section~\ref{sec:data} yields $n=90811$ data entries, $\{ (e_i, y_i \}_{i=1}^n$, extracted from 14854 unique patients.
The dataset is then split 75/25 into a train and a test set.
Importantly, to reduce the risk of leakage between the train and test set, we enforce the data pertaining to a given patient to only belong to one of the sets.

As outlined in Section~\ref{sec:oae}, to map the embeddings into a diagnosis, we employ an SLM $\theta_{SLM}$ along with one projector module for each source that maps the corresponding embedding into the token space of the SLM.
Herein, we consider two recent SLMs: Gemma-2B~\cite{gemmateam2024gemma} and Phi-3-mini-4k~\cite{abdin2024phi} with 2 and 3.8 billion parameters, respectively.
Moreover, Gemma-2B has a token size of 2048 whereas Phi-3-mini-4k has a token size of 3072.
Hence, for the embeddings of the different sources to map into the token space, they must first be projected into a higher-dimensional space.
Notably, we do not train the SLM but only the projectors.

Let the different sources $\{\mathrm{xr}, \mathrm{axr}, \mathrm{proc}, \mathrm{lab}, \mathrm{chart}, \mathrm{txt} \}$ be indexed from 1 to 6.
To train the projector modules, the embedding of each source is routed via its own projection module and mapped into token space, i.e., $t^i=\theta_{\mathrm{enc}}(e^i)$, $i\in[6]$.
The projection module corresponds to a regularized over-complete autoencoder, see Section~\ref{sec:oae}, with an encoder and a decoder realized by one-layer MLPs.
To generate disease predictions $\phi$, the tokens are fed into the SLM either individually, i.e., $\phi=\theta_{\mathrm{SLM}}(t^i)$ for $i\in[6]$, or jointly, i.e., $\phi=\theta_{\mathrm{SLM}}(t^1, t^2 \dots, t^6)$, for isolated and joint training, respectively.
The projector modules involved in a given prediction are then updated from the loss defined in~\eqref{eq:loss_both}.

For the training, we employ ADAM~\cite{kingma2014adam} with a batch size of 32, learning rate $5\times 10^{-4}$, and weight decay $3\times 10^{-4}$, for 50 epochs.
To trade between reconstruction and classification errors in~\eqref{eq:loss_both}, we use the L2-norm for $L_{\mathrm{rec}}$ and $\beta=10$.
Furthermore, for the asymmetric loss in~\eqref{eq:asl}, after fine-tuning, we settled for $m=0.05$ and $\gamma_-=4$.

\subsection{Benchmark}

To compare our results to a benchmark on multimodal disease prediction, we draw inspiration from the holistic AI in medicine framework (HAIM) from~\cite{soenksen2022integrated}.
In~\cite{soenksen2022integrated}, the data was generated from 11 different sources, containing four different modalities, and evaluated seven different machine learning architectures, e.g., XGboost~\cite{xgboost} and attentive tabular networks~\cite{arik2021tabnet}, on the same 12 tasks as presented in~\ref{sec:data}.
However, in contrast to this work, the models in~\cite{soenksen2022integrated} are tailored towards a single task.
In total, 14324 models were trained and compared with XGboost consistently demonstrating the strongest performance.
Hence, we compare our results to an XGboost model trained on the data corresponding to a specific task.

\subsection{Results}
\label{sec:results}
From the methodology described in Section~\ref{sec:oae}, we present the performance of six different models together with the benchmark in Table~II. 
We consider two distinct loss functions, ASL and AVG, and two separate language models, Gemma-2B (G) and Phi-3 (Phi3). 
The four different combinations of loss functions and SLMs is presented where the projectors are trained jointly. 
We also include G-ASL iso, where the projectors are trained in isolation, as well as the best single source performance (BSS), representing the top-performing source for each task (obtained by evaluating the predictions from a single modality using the projectors from the isolated training). 
Importantly, the metrics we consider are precision and recall for each task individually. This patient-centric approach emphasizes the performance of the positive class, as it corresponds to identifying patients in need of care.

\textbf{AVG vs ASL:} We observe that both G-AVG and Phi3-AVG has a slightly higher recall score in four of the pathology diagnosis tasks compared to using ASL. 
For the tasks pertaining to length-of-stay and 48h mortality prediction, both AVG models get higher recall than the ones trained using ASL. Since these tasks have the largest number of data points, these labels are always included in the batch, leading to a large impact on the loss when losses from all labels are averaged. 
This could simultaneously lead to a decrease in performance for some of the labels with fewer data points, hence, explaining why the models using ASL outperforms the AVG models in several tasks. 
Since we consider all labels equally important, a sacrifice in performance for length-of-stay-prediction and 48h mortality-prediction may be warranted to get a more even performance across tasks.

\textbf{Gemma-2B vs Phi3-mini-4k:} Comparing the two different SLMs, it is evident that Phi3-ASL has a lower precision for many of the tasks compared to G-ASL, while having a higher recall across all tasks. 
Projecting into the different token sizes of the SLMs yields comparable results and demonstrates that the method is model agnostic. 
Notably, projecting into higher dimensional spaces risks diluting the signal, and, for this reason, the Phi-3-mini-4k may not correctly classify the negatives as well Gemma-2B. 
However, Phi-3-mini-4k is also the larger model of the two, and the increase in number of parameters could lead to improvement in performance for the prioritized positive class as seen in the higher recall scores. 
Further experiments using other SLMs of different sizes would be needed to draw definite conclusions.

\textbf{Joint vs isolated training:}
We observe a notable improvement when projectors are trained jointly as opposed to isolated training. A comparison between G-ASL and G-ASL iso reveals that, although they yield similar precision scores, G-ASL exhibits significantly better recall. This indicates that letting the SLM reason across the sources and modalities during training is advantageous. Similarly, one can compare G-ASL with BSS. BSS demonstrates behaviour similar to G-ASL iso, with comparable precision but consistently worse recall. Interestingly, when comparing G-ASL iso with BSS, BSS achieves both higher precision and recall in 9 out of 12 tasks, suggesting that training projectors in isolation but using them jointly during prediction could harm the model. A possible explanation for this could reside in the training process. When projectors are trained in isolation, the loss for each projector depends solely on its individual performance. When all projectors are then used jointly, the performance could worsen, as this setting does not align with the training environment.

\textbf{Comparison to Benchmark:} When comparing the results against our benchmark, we observe that many proposed models are on par with or outperform the benchmark for certain lung pathogens. This is particurarly evident for pneumothorax, where all jointly trained models demonstrate better recall. 
However, the XGBoost benchmark is superior on some tasks, as evident from examining the performance on length of stay-prediction and 48 mortality-prediction. 
While recall for G- and Phi3- on these tasks are somewhat competitive, the discrepancy in precision is significant. XGBoost's better precision is also consistent for the lung pathogens, where the difference is more apparent in tasks with few positives. 

Considering that none of the proposed models are getting the same performance as XGBoost, it suggests that the projectors may not effectively project the information to the SLM. 
Another option is to re-evaluate which modalities and sources should be projected and which ones can be adequately understood as text by the SLM. Referring back to Section~\ref{sec:timeseries}, we introduce the time series features as mean, minimum, maximum, etc., per event. Representing them as text might be more effective than as a feature vector projected to the embedding space. Nonetheless, the chosen approach was selected to maintain consistency and one-to-one comparisons to the HAIM framework~\cite{soenksen2022integrated}.
Yet another possible explanation for the performance gap could be the difference in objectives. The benchmark employs a single XGBoost model per task, whereas our models are trained to predict multiple labels simultaneously.

\begin{table*}[t]
\label{tab:results}
\caption{Precision and recall scores for each model and task. The models and methods used are: G-ASL (Gemma-2B with asymmetric loss), G-AVG (Gemma-2B with averaged binary cross-entropy loss), Phi3-ASL (Phi3-mini-4k with asymmetric loss), PHI3-AVG (Phi3-mini-4k with averaged binary cross-entropy loss), G-ASL iso (G-ASL with isolated projector training), and BSS (best single source performance).}
\centering
\begin{tabular}{|c|*{14}{c|}}
\hline
\textbf{Task} & 
\multicolumn{2}{c|}{\textbf{HAIM~\cite{soenksen2022integrated}}} & 
\multicolumn{2}{c|}{\textbf{G-ASL}} & 
\multicolumn{2}{c|}{\textbf{G-AVG}} &
\multicolumn{2}{c|}{\textbf{Phi3-ASL}} &
\multicolumn{2}{c|}{\textbf{Phi3-AVG}} &
\multicolumn{2}{c|}{\textbf{G-ASL iso}} &
\multicolumn{2}{c|}{\textbf{BSS}} \\ \hline
 & Prec & Rec & Prec & Rec & Prec & Rec & Prec & Rec & Prec & Rec & Prec & Rec & Prec & Rec \\ \hline
Fracture & 0.958 & 0.999 & 0.938 & 0.944 & 0.944 & 0.932 & 0.938 & 0.942 & 0.937 & 0.926 & 0.945 & 0.91 & 0.943 & 0.712 \\
Lung Lesion & 0.949 & 0.998 & 0.939 & 0.965 & 0.938 & 0.992 & 0.938 & 0.997 & 0.938 & 0.995 & 0.946 & 0.76 & 0.953 & 0.875 \\
Enlarged CM & 0.866 & 0.94 & 0.782 & 0.903 & 0.795 & 0.868 & 0.793 & 0.95 & 0.774 & 0.962 & 0.756 & 0.924 & 0.8 & 0.723 \\
Consolidation & 0.915 & 0.976 & 0.86 & 0.974 & 0.866 & 0.974 & 0.869 & 0.986 & 0.855 & 0.991 & 0.86 & 0.908 & 0.882 & 0.922 \\
Pneumonia & 0.831 & 0.852 & 0.663 & 0.828 & 0.672 & 0.817 & 0.631 & 0.934 & 0.669 & 0.802 & 0.639 & 0.391 & 0.648 & 0.717 \\
Atelectasis & 0.982 & 0.999 & 0.976 & 0.991 & 0.976 & 0.995 & 0.976 & 0.999 & 0.976 & 0.999 & 0.976 & 0.824 & 0.977 & 0.935 \\
Lung Opacity & 0.973 & 0.999 & 0.964 & 0.985 & 0.964 & 0.996 & 0.964 & 0.997 & 0.964 & 0.999 & 0.966 & 0.878 & 0.967 & 0.941 \\
Pneumothorax & 0.879 & 0.461 & 0.297 & 0.53 & 0.34 & 0.567 & 0.289 & 0.772 & 0.322 & 0.566 & 0.205 & 0.34 & 0.328 & 0.446 \\
Edema & 0.876 & 0.905 & 0.699 & 0.89 & 0.725 & 0.868 & 0.71 & 0.958 & 0.71 & 0.907 & 0.722 & 0.691 & 0.727 & 0.817 \\
Cardiomegaly & 0.906 & 0.96 & 0.824 & 0.966 & 0.826 & 0.954 & 0.828 & 0.985 & 0.82 & 0.989 & 0.833 & 0.832 & 0.842 & 0.894 \\
Length of stay & 0.314 & 0.742 & 0.131 & 0.4 & 0.145 & 0.559 & 0.171 & 0.446 & 0.154 & 0.569 & 0.124 & 0.321 & 0.13 & 0.464 \\
48h Mortality & 0.248 & 0.481 & 0.04 & 0.164 & 0.03 & 0.552 & 0.05 & 0.235 & 0.033 & 0.591 & 0.016 & 0.109 & 0.03 & 0.53 \\
\hline
\end{tabular}
\end{table*}

\section{Conclusion and Future Work}
This paper explores multimodal language models in healthcare, raising the question whether a language model can reason across different modalities to perform better than using a single source. 
To answer this, a joint-fusion model-agnostic framework was presented.
Our framework relies on projecting different data sources, and their corresponding modalities, into the token-space of the language model while ensuring the input signal is maintained in the projection. 
By freezing the weights of the language model, the projectors were trained to map multiple-modalities into token-space to perform multi-disease risk prediction simultaneously across 12 different tasks. 
Our results demonstrate that multimodal models outperformed single-source models in the majority of tasks. 
Additionally, the streamlined approach of using a single model, as opposed to having one XGBoost model per task, demonstrated the multimodal approach's ability to generalize well across various modalities, sources and tasks. 
This work aims to inspire further improvements in the field as a step towards a holistic disease risk prediction, and ultimately medical assistants, utilizing the generative capabilities of SLMs together with reliable classification of multimodal inputs. Moreover, by refining our approach and by incorporating instruction fine-tuning of the SLM, we believe there is potential for significant improvements.

\bibliographystyle{ieeetr}
\bibliography{references}

\end{document}